\documentclass{article}

\usepackage{nips13submit_e,times}
\usepackage{booktabs}
\usepackage{epsfig}
\usepackage{graphicx}
\usepackage{verbatim}
\usepackage{amsmath}
\usepackage{textcomp}
\usepackage{amssymb}
\usepackage{algorithm}
\usepackage{algorithmic}

\usepackage{amsmath,amsfonts,amssymb}
\usepackage{algorithm}
\usepackage{algorithmic}
\usepackage{multirow}
\usepackage{chngpage}

\usepackage{marginnote}
\usepackage{paralist}
\usepackage[normalem]{ulem}
\usepackage{sidecap}
\usepackage{float}
\usepackage{caption3}
\DeclareCaptionOption{parskip}[]{}
\usepackage[small]{caption}
\newfloat{algorithm}{t}{lop}
\usepackage{rotating}
\usepackage{subfigure}

\usepackage{hyperref}
\usepackage{xcolor}
\definecolor{dark-red}{rgb}{0.4,0.15,0.15}
\definecolor{dark-blue}{rgb}{0.15,0.15,0.4}
\definecolor{medium-blue}{rgb}{0,0,0.5}
\hypersetup{
    colorlinks, linkcolor={black},
    citecolor={dark-blue}, urlcolor={medium-blue}
}

\newcommand{\MMD}{\operatorname{MMD}}
\newcommand{\ra}[1]{\renewcommand{\arraystretch}{#1}}

\graphicspath{{../}}

\usepackage{wrapfig}

\title{$B$-tests: Low Variance Kernel Two-Sample Tests}

\author{Wojciech Zaremba\\
Center for Visual Computing\\ 
\'Ecole Centrale Paris\\
Ch\^{a}tenay-Malabry, France\\
\And
Arthur Gretton\\
Gatsby Unit\\
University College London\\
United Kingdom\\
\And
Matthew Blaschko\\
\'{E}quipe GALEN\\ 
Inria Saclay\\
Ch\^{a}tenay-Malabry, France\\
\and
\texttt {\{woj.zaremba,arthur.gretton\}@gmail.com, matthew.blaschko@inria.fr}
}

\newcommand{\BigO}[1]{\ensuremath{\operatorname{O}\bigl(#1\bigr)}}

\nipsfinalcopy 

\begin{document}

\maketitle

\newtheorem{theorem}{Theorem}[section]
\newtheorem{lemma}[theorem]{Lemma}
\newtheorem{proposition}[theorem]{Proposition}
\newtheorem{corollary}[theorem]{Corollary}

\definecolor{LimeGreen}{rgb}{0.1,0.9,0.1}
\definecolor{Maroon}{rgb}{0.9,0.1,0.1}

\begin{abstract}

A family of maximum mean discrepancy (MMD) kernel two-sample tests is introduced.  Members of the test family are called Block-tests or $B$-tests, since  the test statistic is an average over MMDs computed on subsets of the samples. The choice of block size allows control over the tradeoff between test power and computation time. 
In this respect, the $B$-test family combines favorable properties of previously proposed $\MMD$ two-sample tests: $B$-tests are more powerful than a linear time test where blocks are just pairs of samples, yet they are more computationally efficient than a quadratic time test where a single large block incorporating all the samples is used to compute a U-statistic. A further important advantage of the $B$-tests is their asymptotically Normal null distribution: this is by contrast with the U-statistic, which is degenerate under the null hypothesis, and for which estimates of the null distribution are computationally demanding.
Recent results on kernel selection for hypothesis testing  transfer seamlessly to the $B$-tests, yielding a means to optimize test power via kernel choice.

\end{abstract}

\vspace{-0.3cm}
\section{Introduction}
\vspace{-0.3cm}

Given two samples $\{x_{i}\}_{i=1}^{n}$ where $x_{i}\sim P$ i.i.d.,
and $\{y_{i}\}_{i=1}^{n}$, where $y_{i}\sim Q$ i.i.d, the two sample
problem consists in testing whether to accept or reject the null hypothesis
$\mathcal{H}_0$ that $P=Q$, vs the alternative hypothesis $\mathcal{H}_A$ that 
$P$ and $Q$ are different. 
This problem has recently been addressed 
using  measures
of similarity computed in a reproducing kernel Hilbert space (RKHS), which apply
in very general settings where $P$ and $Q$ might
be distributions over high dimensional data or structured objects.
Kernel test statistics  include  the maximum mean discrepancy \cite{Gretton2012KTT,FroLauLerRey12}
(of which the energy distance is an example \cite{SejGreSriFuk12,Baringhaus2004,Szekely2004}),
which is the distance between expected features of $P$ and $Q$ in the RKHS;
the kernel Fisher discriminant \cite{HarBacMou08}, which is the
distance between expected feature maps normalized by the feature space covariance; and density ratio
estimates \cite{YamSuzKanHAcetal13}. When used in testing, it is
necessary to determine whether the empirical estimate of the relevant
similarity measure is sufficiently large as to give the hypothesis
$P=Q$ low probability; i.e., below a user-defined threshold $\alpha$,
denoted the test level. The test power denotes the probability of correctly rejecting
the null hypothesis, given that $P\neq Q$.

The minimum variance unbiased estimator $\mathrm{MMD}_{u}$ of the
maximum mean discrepancy, on the basis of $n$ samples observed from
each of $P$ and $Q$, is a U-statistic, costing $O(n^{2})$ to compute.
Unfortunately, this statistic is degenerate under the null hypothesis
$\mathcal{H}_{0}$ that $P=Q$, and its asymptotic distribution
takes the form of an infinite weighted sum of independent $\chi^{2}$
variables (it is asymptotically Gaussian under the alternative hypothesis
$\mathcal{H}_{A}$ that $P\neq Q$). Two methods for empirically estimating
the null distribution in a consistent way have been proposed: the
bootstrap \cite{Gretton2012KTT}, and a method requiring an eigendecomposition
of the kernel matrices computed on the merged samples from $P$ and $Q$ \cite{afastconsistent}.
Unfortunately, both procedures are computationally demanding: the
former costs $O(n^{2})$, with a large constant (the MMD must be computed
repeatedly over random assignments of the pooled data); the latter
costs $O(n^{3})$, but with a smaller constant, hence can in practice
be faster than the bootstrap. Another approach is to approximate the
null distribution by a member of a simpler parametric family (for
instance, a Pearson curve approximation), however this has no consistency
guarantees. 

More recently, an $O(n)$ unbiased estimate $\mathrm{MMD}_{l}$ of the maximum
mean discrepancy has been proposed \cite[Section 6]{Gretton2012KTT},
which is simply a running average over independent pairs of samples
from $P$ and $Q$. While this has much greater variance than the
U-statistic, it also has a simpler null distribution: being an average
over i.i.d. terms, the central limit theorem gives an asymptotically
Normal distribution, under both $\mathcal{H}_{0}$ and $\mathcal{H}_{A}$.
It is shown in \cite{GrettonSSSBPF2011} that this simple asymptotic
distribution makes it easy to optimize the Hodges and Lehmann asymptotic
relative efficiency \cite{Serfling80} over the family of kernels
that define the statistic: in other words, to choose the kernel which
gives the lowest Type II error (probability of wrongly accepting $\mathcal{H}_0$) for a given Type I error (probability of wrongly rejecting $\mathcal{H}_0$). Kernel selection
for the U-statistic is a much harder question due to the complex form
of the null distribution, and remains an open problem. 

It appears that $\mathrm{MMD}_{u}$ and $\mathrm{MMD}_{l}$ fall at
two extremes of a spectrum: the former has the lowest variance of
any $n$-sample estimator, and should be used in limited data regimes;
the latter is the estimator requiring the least computation while
still looking at each of the samples, and usually achieves better
Type II error than $\mathrm{MMD}_{u}$ at a given computational cost,
albeit by looking at much more data (the ``limited time, unlimited
data'' scenario). A major reason $\mathrm{MMD}_{l}$ is faster is
that its null distribution is straightforward to compute, since it is Gaussian
 and its variance can be calculated at the same cost as the test statistic.
A reasonable next step would be to find a compromise between these
two extremes: to construct a statistic with a lower variance than
$\mathrm{MMD}_{l}$, while retaining an asymptotically Gaussian null
distribution (hence remaining faster than tests based on $\mathrm{MMD}_{u}$). 
We study a family of such test statistics, where we
split the data into blocks of size $B$, compute the quadratic-time $\mathrm{MMD}_{u}$
on each block, and then average the resulting statistics. We call
the resulting tests $B$-tests. 
 As long as we choose the size $B$
of blocks such that $n/B\rightarrow\infty$, we are still guaranteed
asymptotic Normality by the central limit theorem, and the null distribution
can be computed at the same cost as the test statistic. %
For a given
sample size $n$, however, the power of the test can increase dramatically
over the $\mathrm{MMD}_{l}$ test, even for moderate block sizes $B$,
making much better use of the available data with only a small increase
in computation.

The block averaging scheme was originally proposed in
\cite{HoShieh06}, as an instance of a two-stage U-statistic, to be applied 
when the degree of degeneracy of the U-statistic is indeterminate.
Differences with respect to our method are that Ho and Shieh
compute the block statistics by sampling with replacement
\cite[(b) p. 863]{HoShieh06}, and propose to obtain the variance 
of the test statistic via Monte Carlo, jackknife, or bootstrap techniques,
whereas we use closed form expressions. Ho and Shieh
further suggest an alternative two-stage U-statistic in the event that the
degree of degeneracy is known; we return to this point in the discussion.
While we confine ourselves to the MMD in this paper, we emphasize
that the block approach applies to a much broader variety of test situations where the
null distribution cannot easily be computed, including the energy distance
and distance covariance \cite{SejGreSriFuk12,Baringhaus2004,Szekely2004}
 and Fisher statistic \cite{HarBacMou08} in the case
of two-sample testing, and the Hilbert-Schmidt Independence Criterion
\cite{GreFukTeoSonetal08} and distance covariance \cite{SzeRizBak07} for independence testing.
 Finally, the kernel learning approach of \cite{GrettonSSSBPF2011}
applies straightforwardly, allowing us to maximize test power over
a given kernel family.
Code 
is available at \url{http://github.com/wojzaremba/btest}.

\begin{figure}
  \centering
  \subfigure[$B=2$. This setting corresponds to the $\MMD_l$ statistic~\cite{Gretton2012KTT}.]{
	\includegraphics[width=0.47\textwidth]{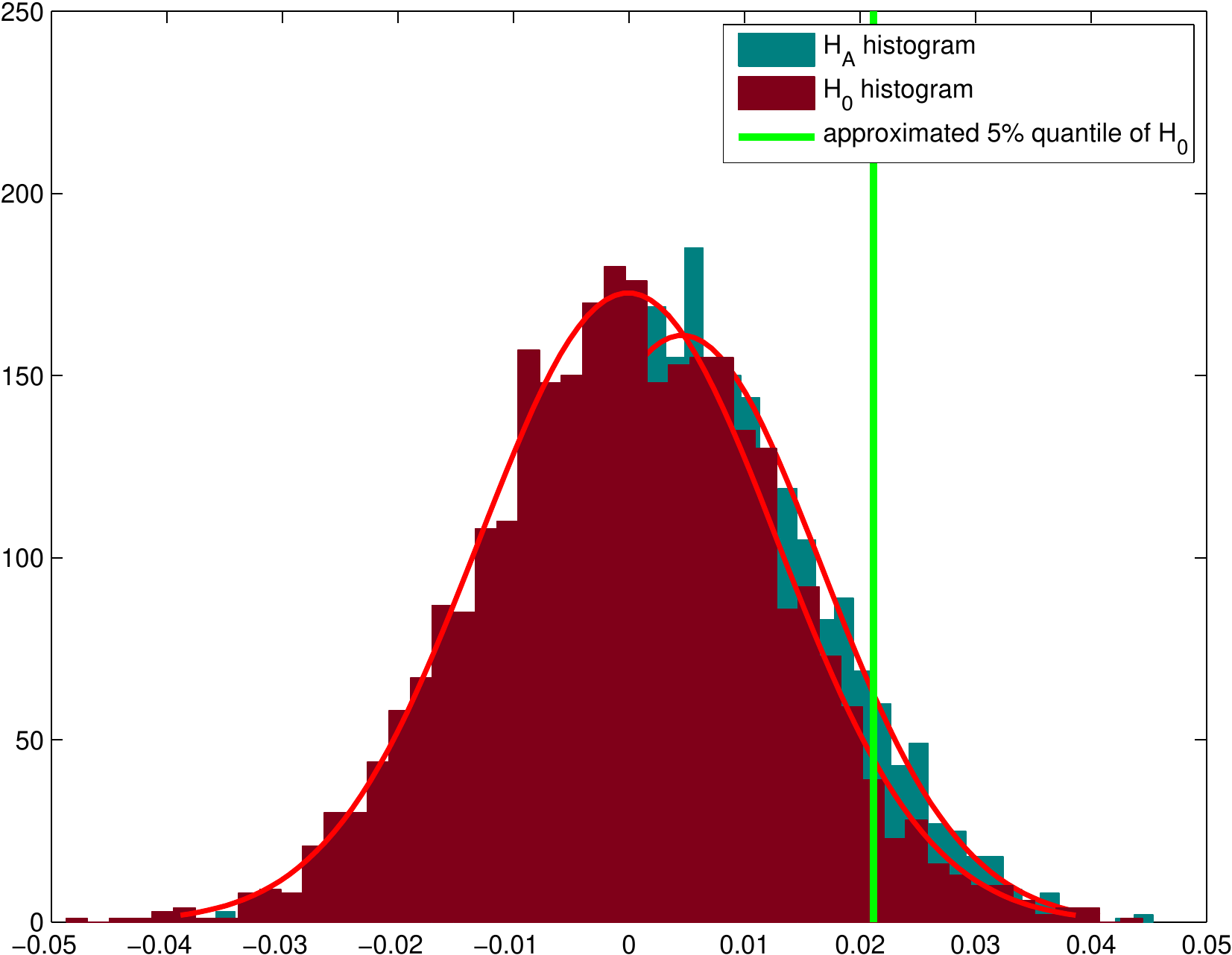}
  }
\hfill
  \subfigure[$B=250$ ]{
	\includegraphics[width=0.47\textwidth]{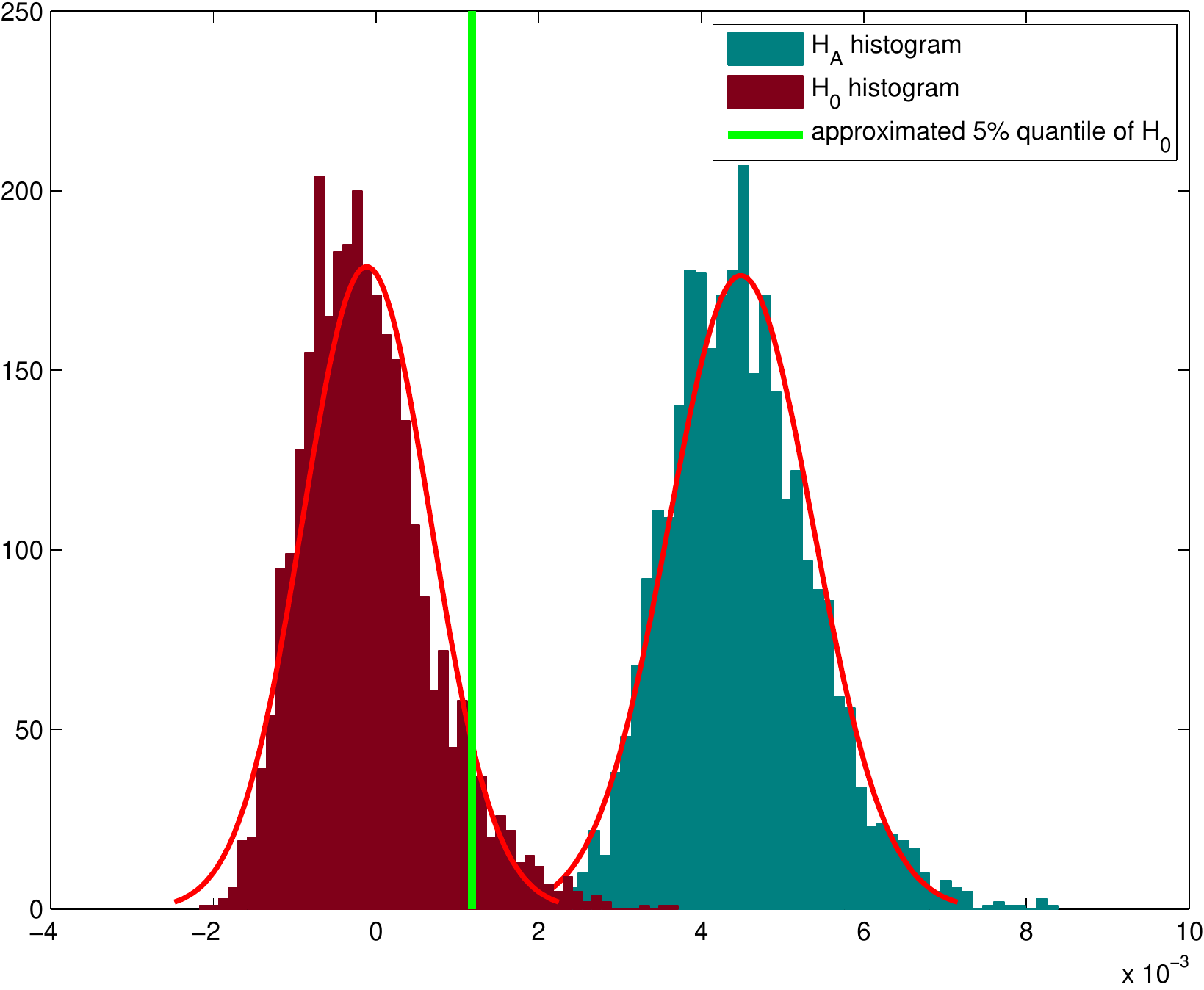}
  }
  \caption{
Empirical distributions under $\mathcal{H}_0$ and $\mathcal{H}_A$ for different regimes of $B$ for the music experiment (Section~\ref{sec:MusicExperiment}).  In both plots, the number of samples is fixed at 500.
  As we vary $B$, we trade off the quality of the finite sample Gaussian approximation to the null distribution, as in Theorem~\ref{theorem:berry}, with the variances of the $\mathcal{H}_0$ and $\mathcal{H}_A$ distributions, as outlined in Section~\ref{sec:AsymptoticsConsistency}. 
In (b) the distribution under $\mathcal{H}_0$ does not resemble a Gaussian (it does not pass a level $0.05$ Kolmogorov-Smirnov (KS) normality test~\cite{kolmogorov1933sulla,smirnov1948table}), and 
a Gaussian approximation results in a conservative test threshold (vertical green line).  The remaining empirical distributions all pass a KS normality test.}
  \label{fig:EmpiricalDistsLargeSmallK}
\end{figure}

\vspace{-0.3cm}
\section{Theory}\label{sec:theory}
\vspace{-0.3cm}

In this section we describe the mathematical foundations of the $B$-test. 
We begin with a brief review of kernel methods, and of the maximum
mean discrepancy.
We then present our block-based average MMD statistic, and derive its distribution under the $\mathcal{H}_0$ ($P = Q$) and $\mathcal{H}_A$ ($P \neq Q$) hypotheses.  The central idea employed in the construction of the $B$-test is to generate 
a low variance MMD estimate by averaging multiple low variance kernel statistics computed over blocks of samples.  We show simple sufficient conditions on the block size for consistency of the estimator.  Furthermore, we analyze the properties of the finite sample estimate, and propose a consistent strategy for setting the block size as a function of the number of samples.

\subsection{Definition and asymptotics of the block-MMD}\label{sec:AsymptoticsConsistency}

Let $\mathcal{F}_{k}$ be an RKHS
defined on a topological space $\mathcal{X}$ with reproducing kernel
$k$, and $P$ a Borel probability measure on $\mathcal{X}$. The
\emph{mean embedding} of $P$ in $\mathcal{F}_{k}$, written $\mu_{k}(p)\in\mathcal{F}_{k}$
is defined such that $E_{x\sim p}f(x)=\left\langle f,\mu_{k}(p)\right\rangle _{\mathcal{F}_{k}}$
for all $f\in\mathcal{F}_{k}$, and exists for all Borel probability
measures when $k$ is bounded and continuous \cite{BerTho04,Gretton2012KTT}. The
maximum mean discrepancy (MMD) between a Borel probability measure
$P$ and a second Borel probability measure $Q$ is the squared RKHS
distance between their respective mean embeddings,
\begin{equation}
\eta_{k}(P,Q)=\left\Vert \mu_{k}(P)-\mu_{k}(Q)\right\Vert _{\mathcal{F}_{k}}^{2}=E_{xx'}k(x,x')+E_{yy'}k(y,y')-2E_{xy}k(x,y),\label{eq:MMD}
\end{equation}
where $x'$ denotes an independent copy of $x$ \cite{GrettonBRSS06}.
Introducing the notation $z=(x,y)$, we write
\begin{equation}
\eta_{k}(P,Q)=E_{zz'}h_{k}(z,z'),
\qquad
h(z,z')=k(x,x')+k(y,y')-k(x,y')-k(x',y).
\label{eq:MMD_with_h}
\end{equation}
When the kernel $k$ is characteristic, then $\eta_{k}\left(P,Q\right)=0$
iff $P=Q$ \cite{SriGreFukLanetal10}. 
Clearly, the minimum variance 
unbiased estimate $\MMD_u$ of $\eta_{k}(P,Q)$ is a U-statistic.

By analogy with $\MMD_u$, we make use of averages of $h(x,y,x',y')$ to construct our two-sample test.
We denote by $\hat{\eta}_k(i)$ the $i$th empirical estimate $\MMD_u$ based on a subsample of size $B$, where $1 \leq i \leq \frac{n}{B}$ (for notational purposes, we will index samples as though they are presented in a random fixed order).  More precisely,
\begin{equation}
  \hat{\eta}_k(i) = \frac{1}{B(B-1)} \sum_{a={(i-1) B + 1}}^{iB} \sum_{b={(i-1)B + 1}, b \neq a}^{iB} h(z_a, z_b) .
\end{equation}
The $B$-test statistic is an $\MMD$ estimate  obtained by averaging the $\hat{\eta}_k(i)$. Each $\hat{\eta}_k(i)$ under $\mathcal{H}_0$ converges to an infinite sum of weighted $\chi^2$ variables~\cite{afastconsistent}. Although setting $B=n$ would lead to the lowest variance estimate of the MMD, computing sound thresholds for a given $p$-value is expensive, involving repeated bootstrap sampling~\cite{efron93bootstrap,Johnson1994}, or computing the eigenvalues of a Gram matrix~\cite{afastconsistent}.

In contrast, we note that ${\hat{\eta}_k(i)}_{i=1,...,\frac{n}{B}}$ are i.i.d.\ variables, and averaging them allows us to apply the central limit theorem in order to estimate $p$-values from a normal distribution. 
We denote the average of the $\hat{\eta}_k(i)$ by $\hat{\eta}_k$,
\begin{equation}
 \hat{\eta}_k = \frac{B}{n} \sum_{i=1}^{\frac{n}{B}} \hat{\eta}_k(i) .
\end{equation}
We would like to apply the central limit theorem to variables ${\hat{\eta}_k(i)}_{i=1,...,\frac{n}{B}}$. 
It remains for us to derive the distribution of $\hat{\eta}_k$ under $\mathcal{H}_0$ and under $\mathcal{H}_A$.  We rely on the result from \cite[Theorem 8]{GrettonBRSS06} for $\mathcal{H}_A$. According to our notation, for every $i$,

\begin{theorem}
  Assume $0 < \mathbb{E}(h^2) < \infty$, then under $\mathcal{H}_A$, $\hat{\eta}_k$ converges in distribution to a Gaussian according to 
  \begin{align}
    B^{\frac{1}{2}}(\hat{\eta}_k(i) - \MMD^2) \overset{D}{\rightarrow} \mathcal{N}(0, \sigma_u^2) ,
  \end{align}
where $\sigma_u^2 = 4 \left (\mathbb{E}_z[(\mathbb{E}_{z'}h(z, z'))^2 - \mathbb{E}_{z, z'}(h(z, z'))]^2\right)$.\end{theorem}

This in turn implies that
\begin{align}
    \hat{\eta}_k(i) \overset{D}{\rightarrow} \mathcal{N}(\MMD^2, \sigma_u^2 B^{-1}).
\end{align}
For an average of $\{\hat{\eta}_k(i)\}_{i=1,\dots,\frac{n}{B}}$, the central limit theorem implies that under $\mathcal{H}_A$,
\begin{align}
    \hat{\eta}_k \overset{D}{\rightarrow} \mathcal{N}\left(\MMD^2, 
\sigma_u^2 \left( B n/B \right)^{-1}
\right) = \mathcal{N}\left(\MMD^2, \sigma_u^2 n^{-1}\right) . \label{eq:ConvergenceHA}
\end{align}
This result shows that  the distribution of $\mathcal{H}_A$ is asymptotically independent  of the block size, $B$.
%
%
%
Turning to the null hypothesis, \cite[Theorem 8]{GrettonBRSS06} additionally implies that under $\mathcal{H}_0$ for every $i$,
\begin{theorem}
\begin{align}\label{eq:infChi2}
  B \hat{\eta}_k(i) \overset{D}{\rightarrow} \sum_{l=1}^\infty \lambda_l [z_l^2 - 2],
\end{align}
where $z_l \sim \mathcal{N}(0, 2)^2$ i.i.d, $\lambda_l$ are the solutions to the eigenvalue equation
\begin{align}
\int_{\mathcal{X}} \bar{k}(x,x') \psi_l(x) dp(x) = \lambda_l \psi_l(x'),
\end{align}
and $\bar{k}(x_i,x_j) := k(x_i, x_j) - \mathbb{E}_xk(x_i,x)  -\mathbb{E}_xk(x,x_j)  + \mathbb{E}_{x,x'} k(x,x')$ is the centered RKHS kernel.
\end{theorem}

As a consequence, under $\mathcal{H}_0$, $\hat{\eta}_k(i)$ has expected variance $2B^{-2}\sum_{l=1}^\infty \lambda^2$. We will denote this variance by $CB^{-2}$. The central limit theorem implies that under $\mathcal{H}_0$,

\begin{align}
    \hat{\eta}_k \overset{D}{\rightarrow} \mathcal{N}\left(0, 
C \left( B^2 n/B\right)^{-1}
\right) = 
\mathcal{N}\left(0, 
C(nB)^{-1}
\right) \label{eq:ConvergenceH0}
\end{align}

The asymptotic distributions for $\hat{\eta}_k$ under $\mathcal{H}_0$ and $\mathcal{H}_A$ are Gaussian, and consequently it is easy to calculate the distribution quantiles and test thresholds. Asymptotically, it is always beneficial to increase $B$, as the distributions for $\eta$ under $\mathcal{H}_0$ and $\mathcal{H}_A$ will be better separated. For consistency, it is sufficient to ensure that $n/B \rightarrow \infty$.

A related strategy of averaging over data blocks to deal with large sample sizes has recently been developed in   \cite{KleTalSarJor13}, with the goal of efficiently computing bootstrapped estimates of statistics of interest (e.g. quantiles or biases). Briefly, the approach splits the  data (of size $n$) into $s$ subsamples each of size $B$, computes an estimate of the $n$-fold bootstrap on each block, and averages these estimates. The difference with respect to our approach is that we use the asymptotic distribution of the average over block statistics to determine a threshold for a hypothesis test, whereas \cite{KleTalSarJor13} is concerned with proving the consistency of a statistic obtained by averaging over bootstrap estimates on blocks.

\subsection{Convergence of Moments}\label{sec:underestimate}

In this section, we  analyze the convergence of the moments of the $B$-test statistic, and comment on potential sources of bias.

The central limit theorem implies that the empirical mean of $\{\hat{\eta}_k(i)\}_{i=1,\dots,\frac{n}{B}}$ converges to $\mathbb{E}(\hat{\eta}_k(i))$. Moreover it states that the variance $\{\hat{\eta}_k(i)\}_{i=1,\dots,\frac{n}{B}}$ converges to $\mathbb{E}(\hat{\eta}_k(i))^2 - \mathbb{E}(\hat{\eta}_k(i)^2)$. Finally, all remaining moments tend to zero, where the rate of convergence for the $j$th moment is of the order $\left(\frac{n}{B}\right)^{\frac{j + 1}{2}}$~\cite{momentconvergence}. This indicates that the skewness dominates the difference of the distribution from a Gaussian. 

Under both $\mathcal{H}_0$ and $\mathcal{H}_A$, 
thresholds computed from normal distribution tables are asymptotically unbiased.
For finite samples sizes,  however, the bias under  $\mathcal{H}_0$ can be more severe.
From Equation~\eqref{eq:infChi2} we have that under $\mathcal{H}_0$, the summands, $\hat{\eta}_k(i)$, converge in distribution to infinite weighted sums of $\chi^2$ distributions. Every unweighted term of this infinite sum has distribution $\mathcal{N}(0, 2)^2$, which has finite skewness equal to $8$. The skewness for the entire sum is finite and positive,
\begin{equation}
C = \sum_{l=1}^\infty 8 \lambda_l^3 ,
\end{equation}
as $\lambda_l \geq 0$ for all $l$ due to the positive definiteness of the kernel $k$. The skew for the mean of the $\hat{\eta}_k(i)$ converges to $0$ and is positively biased. At smaller sample sizes, test thresholds obtained from the standard Normal table may therefore be inaccurate, as they do not account for this skew. In our experiments, this bias caused the tests to be overly conservative, with lower Type I error than the design level required (Figures~\ref{fig:blobs_emp_type1} and \ref{fig:songs_emp_type1}).


\subsection{Finite Sample Case}

In the finite sample case, we apply the Berry-Ess\'een theorem, which gives conservative bounds on the $\ell_{\infty}$ convergence of a series of finite sample random variables to a Gaussian distribution~\cite{berry1941accuracy}. 
\begin{theorem}\label{theorem:berry}
 Let $X_1, X_2, \dots , X_n$ be i.i.d.\ variables. $\mathbb{E}(X_1) = 0$, $\mathbb{E}(X_1^2) = \sigma^2 > 0$, and $\mathbb{E}(|X_1|^3) = \rho < \infty$.
 Let $F_n$ be a cumulative distribution of $\frac{\sum_{i = 1}^nX_i}{\sqrt{n}\sigma}$, and let $\Phi$ denote the standard normal distribution. Then for every $x$,
\begin{equation}
 |F_n(x) - \Phi(x)| \leq 
C\rho \sigma^{-3}n^{-1/2},
\end{equation}
\noindent
where $C < 1$.
\end{theorem}
This result allows us to ensure fast point-wise convergence of the $B$-test. We have that $\rho(\hat{\eta}_k) = O(1)$, i.e., it is dependent only on the underlying distributions of the samples and not on the sample size. The number of i.i.d.\ samples is $nB^{-1}$. 
Based on Theorem~\ref{theorem:berry}, the point-wise error can be upper bounded by $\frac{O(1)}{O(B^{-1})^{\frac{3}{2}}\sqrt{\frac{n}{B}}} = O(\frac{B^2}{\sqrt{n}})$ under $\mathcal{H}_A$. Under $\mathcal{H}_0$, the error can be bounded by $\frac{O(1)}{O(B^{-2})^{\frac{3}{2}}\sqrt{\frac{n}{B}}} = O(\frac{B^{3.5}}{\sqrt{n}})$.

While the asymptotic results indicate that convergence to an optimal predictor is fastest for larger $B$, the finite sample results support decreasing the size of $B$ in order to have a sufficient number of samples for application of the central limit theorem. As long as $B \rightarrow \infty$ and $\frac{n}{B} \rightarrow \infty$, the assumptions of the $B$-test are fulfilled. 



By varying $B$, we make a fundamental tradeoff in the construction of our two sample test. 
When $B$ is small, we have many samples, hence the null distribution is close to the asymptotic limit provided by the central limit theorem, and the Type I error is estimated accurately. 
The disadvantage of a small $B$ is a lower test power for a given sample size. 
Conversely, if we increase $B$, we will have a lower variance empirical distribution for $\mathcal{H}_0$, hence higher test power, but we may have a poor estimate of 
the number of Type I errors (Figure~\ref{fig:EmpiricalDistsLargeSmallK}).
%
A sensible family of heuristics therefore is to set
\begin{equation}\label{eq:blockSelectionHeuristic}
B=[n^\gamma]
\end{equation}
for some $0 < \gamma < 1$, where we round to the nearest integer.  In this setting the number of samples available for application of the central limit theorem will be $[n^{(1-\gamma)}]$.  For given $\gamma$ computational complexity of the $B$-test is $\BigO{n^{1+\gamma}}$.  We note that any value of $\gamma \in (0,1)$ yields a consistent estimator.  We have chosen $\gamma=\frac{1}{2}$ in the experimental results section, with resulting complexity $\BigO{n^{1.5}}$: we emphasize that this is  a heuristic, and just one choice that fulfils our assumptions.

\section{Experiments}\label{sec:experiments}

We have conducted experiments on challenging synthetic and real datasets in order to empirically measure \begin{inparaenum}[(i)] 
\item sample complexity,
\item computation time, and
\item Type I / Type II errors.
\end{inparaenum}
We evaluate $B$-test performance in comparison to the $\MMD_l$ and $\MMD_u$ estimators, where for the latter
we compare across different strategies for null distribution quantile estimation.


\vspace{-0.2cm}
\subsection{Synthetic data}
\vspace{-0.2cm}

\begin{table*}\centering
\ra{0.9}
\resizebox{\textwidth}{!}{
\begin{tabular}{@{}cccccc@{}}\toprule
\multirow{2}{*}{Method} & \multirow{2}{*}{Kernel parameters} & \multirow{2}{65pt}{\centering Additional parameters} & \multirow{2}{80pt}{\centering Minimum number of samples} & \multirow{2}{60pt}{\centering Computation time (s)} & \multirow{2}{*}{Consistent} \\
&&&&& \\
\cmidrule{1-6}
\multirow{7}{*}{$B$-test} & \multirow{3}{65pt}{\centering $\sigma = 1$} & $B = 2$& 26400 & 0.0012 & $\color{LimeGreen} \checkmark $ \\
&& $B = 8$ &  3850 & 0.0039 & $\color{LimeGreen} \checkmark $ \\
&& $B=\sqrt{n}$ &  886 & 0.0572 & $\color{LimeGreen} \checkmark $ \\
\cmidrule{2-6}
& $\sigma = \text{median}$ & any $B$ & $>60000$ & & $\color{LimeGreen} \checkmark $ \\
\cmidrule{2-6}
& \multirow{3}{65pt}{\centering multiple kernels} & $B = 2$& 37000 & 0.0700 & $\color{LimeGreen} \checkmark $ \\
&& $B = 8$ &  5400 & 0.1295 & $\color{LimeGreen} \checkmark $ \\
&& $B=\sqrt{\frac{n}{2}}$ & 1700 & 0.8332 & $\color{LimeGreen} \checkmark $ \\
\cmidrule{1-4}
Pearson curves & \multirow{4}{65pt}{\centering $\sigma = 1$} & \multirow{8}{65pt}{\centering $B = n$} & 186  & 387.4649 & $\color{Maroon}\times$ \\
Gamma approximation &&  &  183 & 0.2667 & $\color{Maroon}\times$ \\
Gram matrix spectrum && &  186 & 407.3447 & $\color{LimeGreen} \checkmark $ \\
Bootstrap && &  190 & 129.4094 & $\color{LimeGreen} \checkmark $ \\
\cmidrule{1-2}
\cmidrule{4-4}
Pearson curves & \multirow{4}{65pt}{\centering $\sigma = \text{median}$} && \multirow{4}{65pt}{\centering $>60000$, or 2h per iteration timeout} & & $\color{Maroon}\times$ \\
Gamma approximation & & & & & $\color{Maroon}\times$ \\
Gram matrix spectrum & & & & & $\color{LimeGreen} \checkmark $ \\
Bootstrap && & & & $\color{LimeGreen} \checkmark $ \\
\bottomrule
\end{tabular}
}
\caption{Sample complexity for tests on the distributions described in Figure~\ref{fig:blobs}. The fourth column indicates the minimum number of samples necessary to achieve Type I and Type II errors of $5\%$.  The fifth column is the computation time required for 2000 samples, and is not presented for settings that have unsatisfactory sample complexity.  
}
    \label{tab:samples_complexity}
\end{table*}
\begin{figure}
  \centering
  \subfigure[]{\label{fig:blobs_emp_type1a}
    \includegraphics[width=0.31\textwidth]{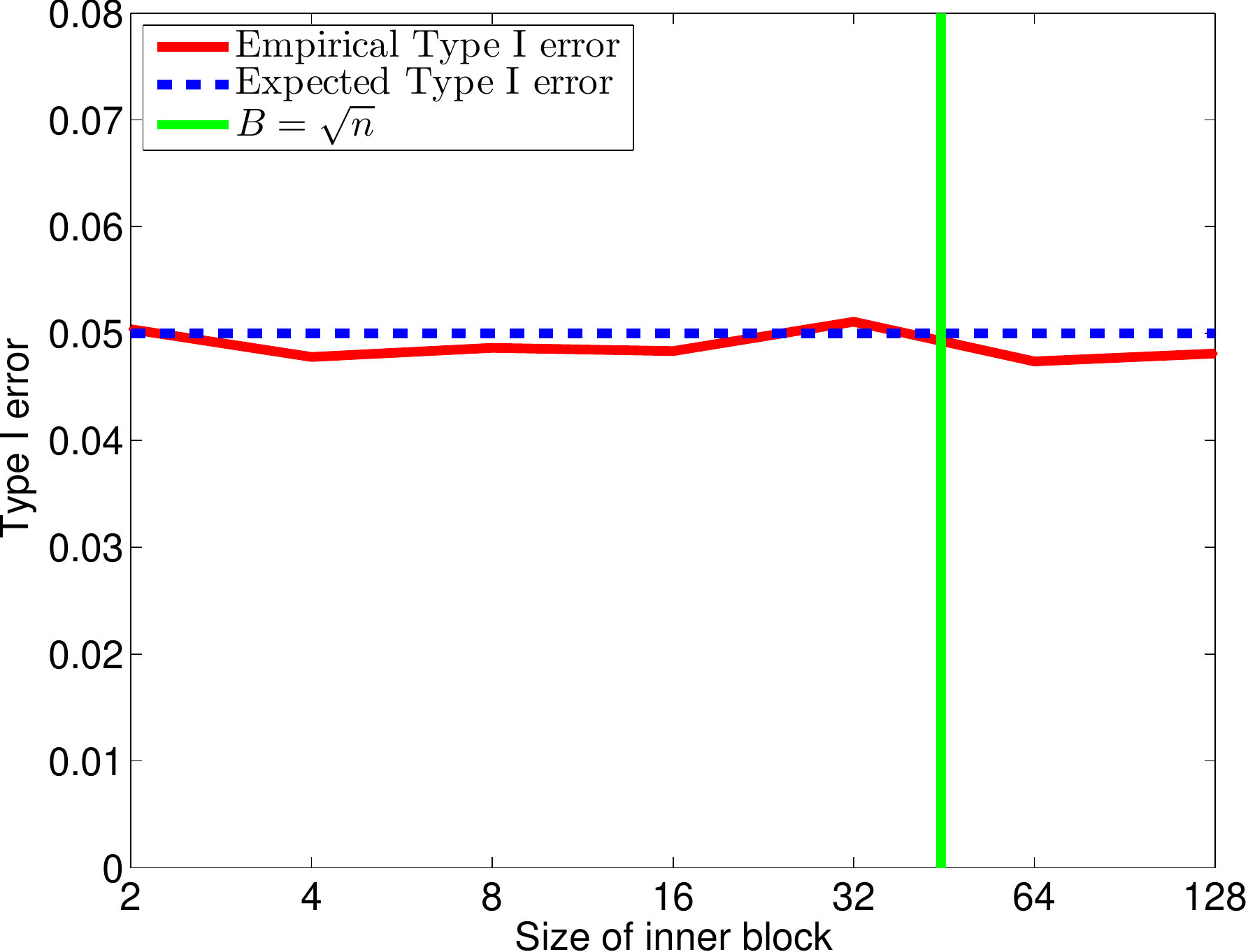}
  }
  \subfigure[]{\label{fig:blobs_emp_type1b}
    \includegraphics[width=0.31\textwidth]{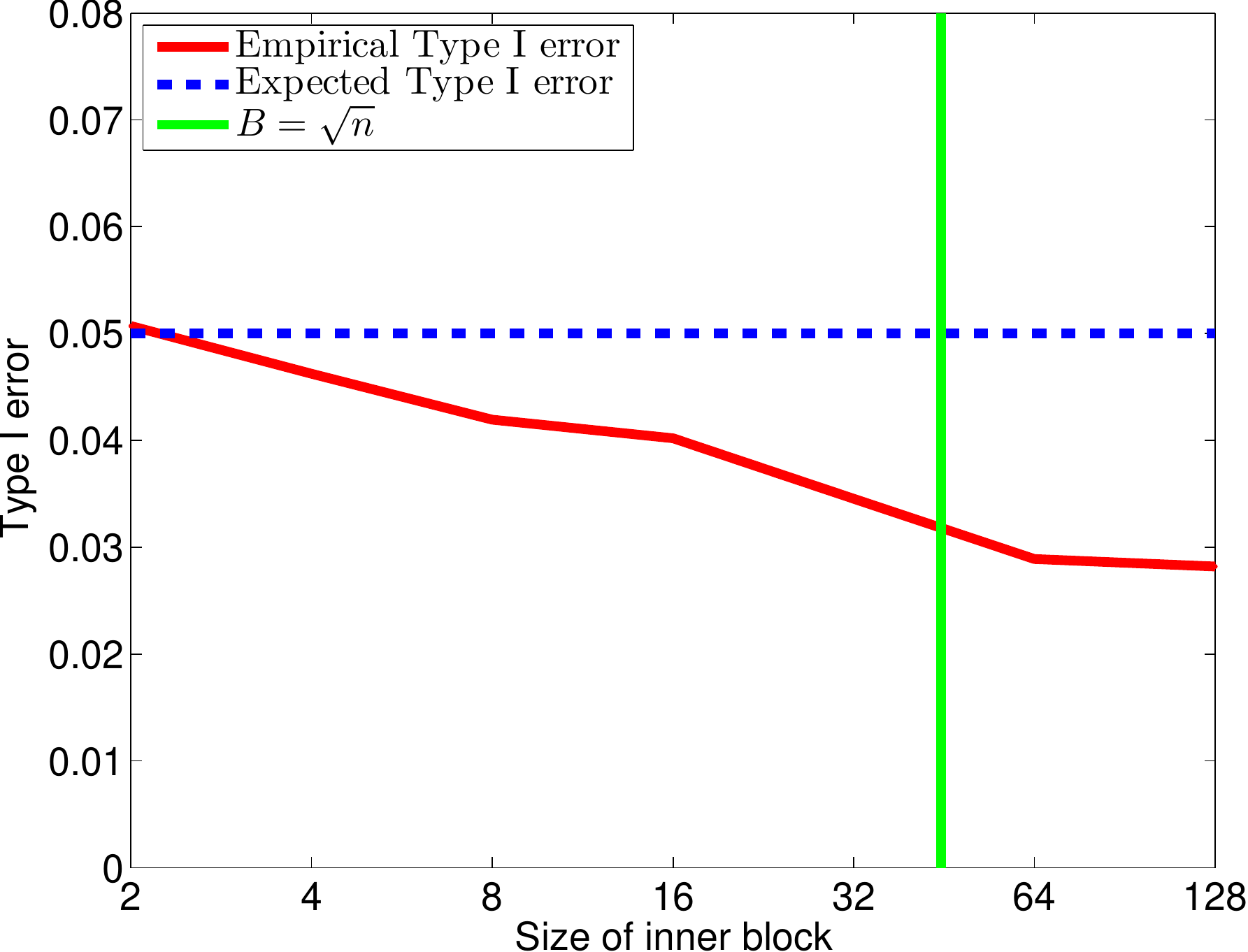}
  }
  \subfigure[]{\label{fig:blobs_emp_type1c}
    \includegraphics[width=0.31\textwidth]{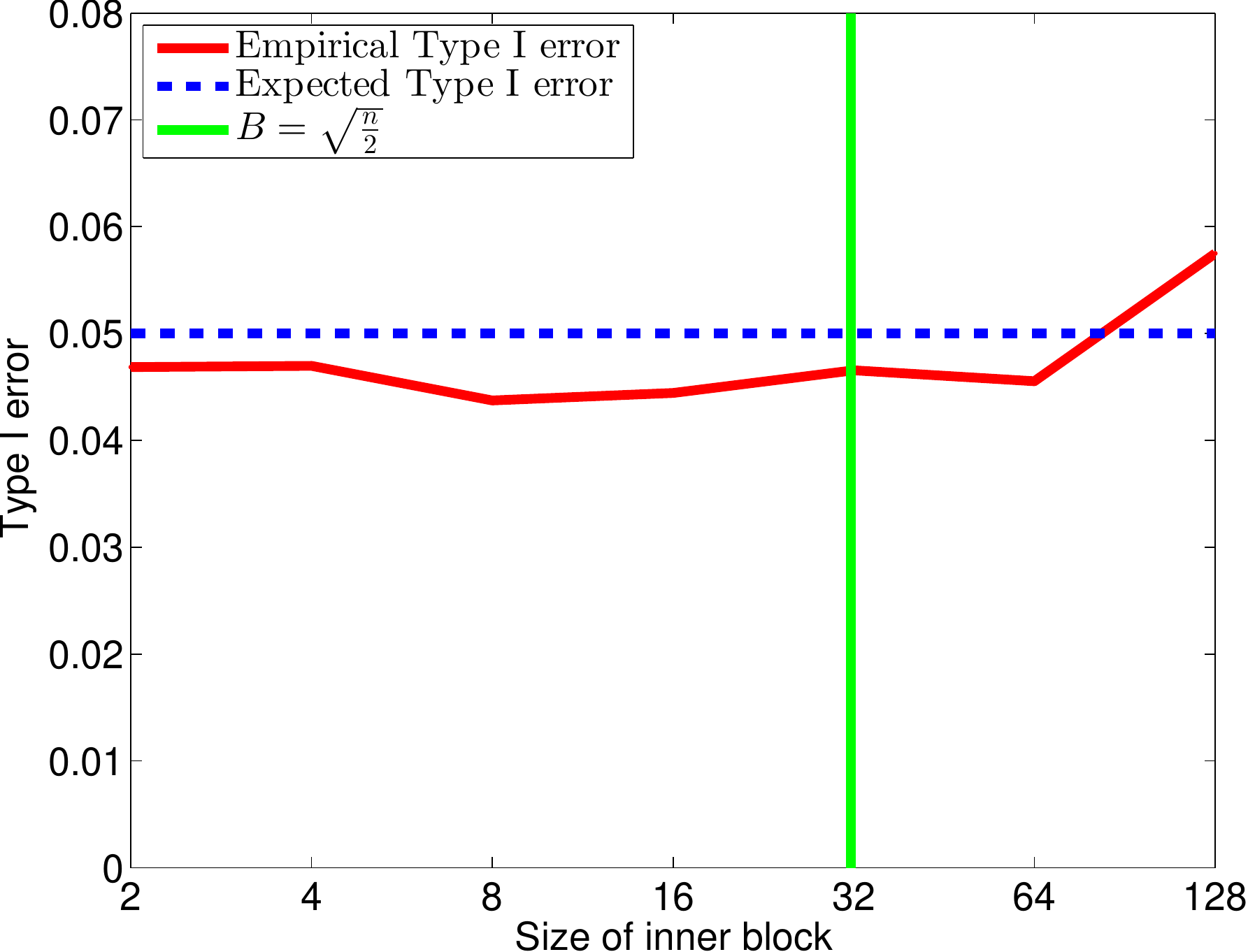}
  }
  \caption{Type I errors on the distributions shown in Figure~\ref{fig:blobs} for $\alpha = 5\%$: (a) MMD, single kernel, $\sigma = 1$, (b) MMD, single kernel, $\sigma$ set to the median pairwise distance, and (c) MMD, non-negative linear combination of multiple kernels. The experiment was repeated $30000$ times.  Error bars are not visible at this scale.}
  \label{fig:blobs_emp_type1}
\vspace{-0.6cm}
\end{figure}

\begin{wrapfigure}{rt}{0.4\textwidth}
\vspace{-0.7cm}
\subfigure[Distribution $P$]{
	\includegraphics[width=0.18\textwidth]{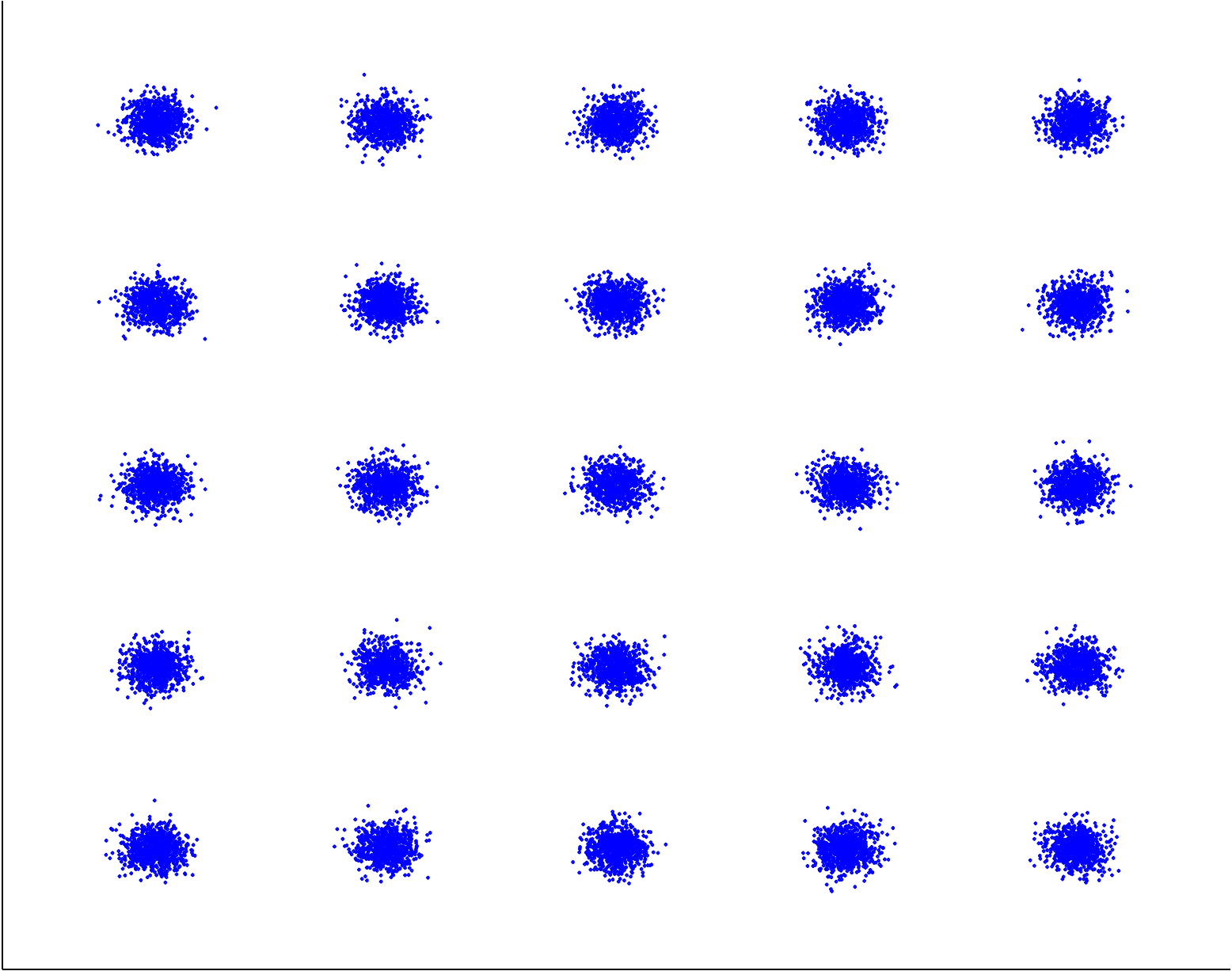}
  }
\hfill
  \subfigure[Distribution $Q$]{
	\includegraphics[width=0.18\textwidth]{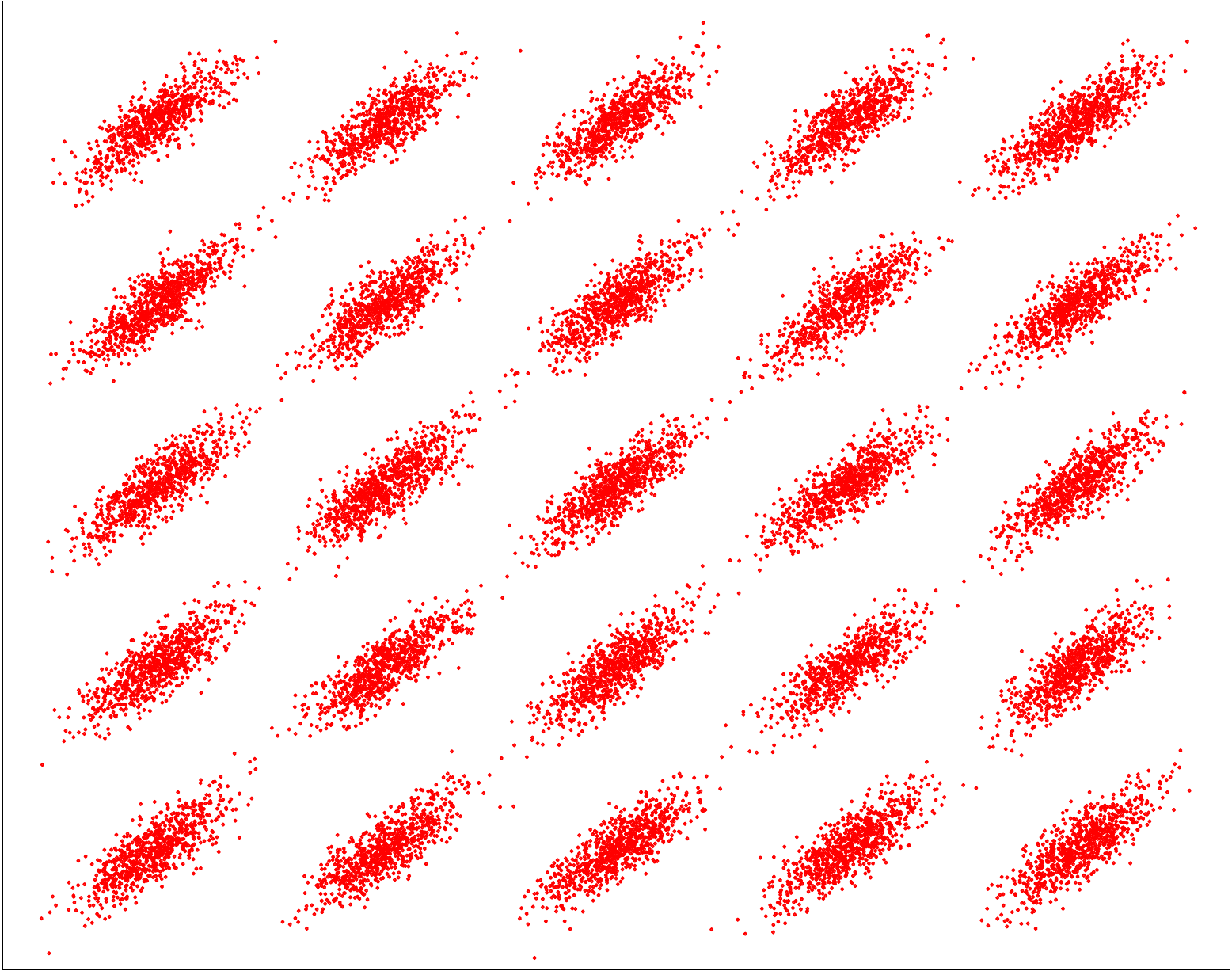}
  }
  \caption{Synthetic data distributions $P$ and $Q$. Samples belonging to these classes are difficult to distinguish.}
  \label{fig:blobs}
\vspace{-0.2cm}
\end{wrapfigure}

Following previous work on kernel hypothesis testing~\cite{GrettonSSSBPF2011}, our synthetic distributions are $5 \times 5$ grids of 2D Gaussians.  We specify two distributions, $P$ and $Q$. For distribution $P$ each Gaussian has identity covariance matrix, while for distribution $Q$ the covariance is non-spherical. Samples drawn from $P$ and $Q$ are presented in Figure~\ref{fig:blobs}.  These distributions have proved to be very challenging for existing non-parametric two-sample tests~\cite{GrettonSSSBPF2011}.

We employed three different kernel selection strategies in the hypothesis test. 
First, we used a Gaussian kernel with $\sigma = 1$, which approximately matches the scale of the variance of each Gaussian in mixture $P$. 
While this is a somewhat arbitrary default choice, we selected it as it performs well in practice (given the lengthscale of the data), and we treat it as a baseline. Next, we set $\sigma$ equal to the median pairwise distance over the training data, which is a standard way to choose the Gaussian kernel bandwidth~\cite{SchoelkopfThesis}, although it is likewise arbitrary in this context. Finally, we applied a kernel learning strategy, in which the kernel was optimized to maximize the test power for the alternative $P \neq Q$~\cite{GrettonSSSBPF2011}. This approach returned a non-negative linear combination combination of base kernels, where half the data were used in learning the kernel weights (these data were excluded from the testing phase).

 The base kernels
 in our experiments were chosen to be Gaussian, with bandwidths in the set $\sigma \in \{2^{-15}, 2^{-14}, \dots, 2^{10}\}$. Testing was conducted using the remaining half of the data.

\begin{wrapfigure}{rt}{0.5\textwidth}
\includegraphics[width=0.48\textwidth]{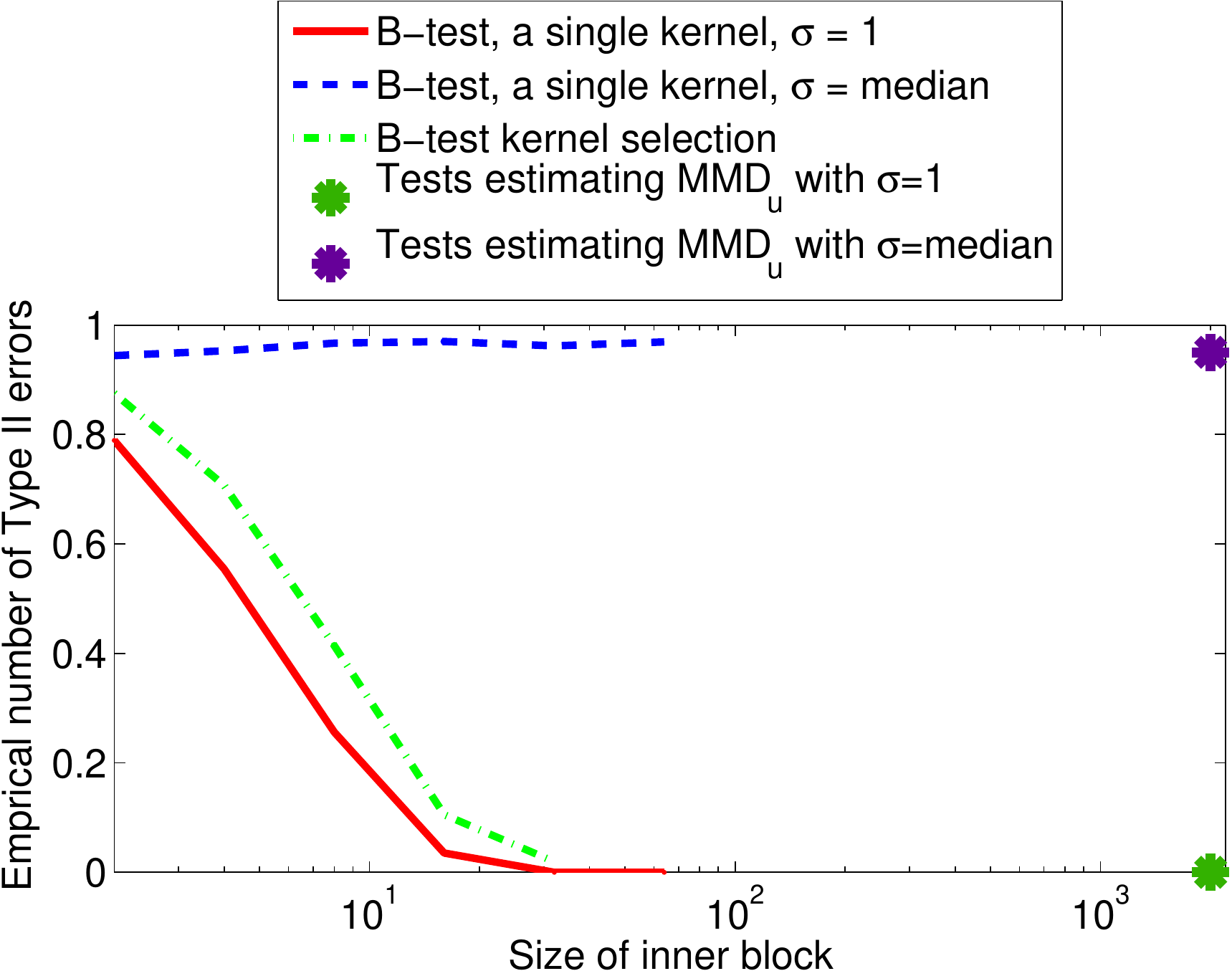}
  \caption{Synthetic experiment: number of Type II errors vs $B$, given a fixed probability $\alpha$ of Type I errors. As $B$ grows, the Type II error drops quickly when the kernel is appropriately chosen.
The kernel selection method is described in~\cite{GrettonSSSBPF2011}, and closely approximates the baseline  performance of the well-informed user choice of $\sigma=1$.}
  \label{fig:blobs_perf}
\vspace{-0.4cm}
\end{wrapfigure}
 For comparison with the quadratic time $U$-statistic $\MMD_u$  \cite{afastconsistent,Gretton2012KTT}, we evaluated four null distribution estimates: \begin{inparaenum}[(i)]
\item Pearson curves, 
\item gamma approximation, 
\item Gram matrix spectrum, and 
\item bootstrap.
\end{inparaenum}
For methods using Pearson curves and the Gram matrix spectrum, we drew $500$ samples from the null distribution estimates to obtain the $1-\alpha$ quantiles, for a test of level $\alpha$. For the bootstrap, we fixed the number of shuffles to $1000$. 
We note that Pearson curves and the gamma approximation are not statistically consistent.
We considered only the setting with $\sigma = 1$ and $\sigma$ set to the median pairwise distance, as kernel selection is not yet solved for tests using $\MMD_u$~\cite{GrettonSSSBPF2011}.

In the first experiment we set the Type I error to be $5\%$, and we recorded the Type II error. We conducted these experiments on $2000$ samples over $1000$ repetitions, with varying block size, $B$. Figure~\ref{fig:blobs_perf} presents results for different kernel choice strategies, as a function of $B$. The median heuristic performs extremely poorly in this experiment. As discussed in \cite[Section 5]{GrettonSSSBPF2011}, the reason for this failure is that the lengthscale of the difference between the distributions $P$ and $Q$ differs from the lengthscale of the main data variation as captured by the median, which gives too broad a kernel for the data.

In the second experiment, our aim was to compare the empirical sample complexity of the various methods. We again fixed  the same Type I error for all methods, but this time we also fixed a Type II error of $5\%$, increasing the number of samples until the latter error rate was achieved. Column four of Table~\ref{tab:samples_complexity} shows the number of samples required in each setting to achieve these error rates.
We additionally compared the computational efficiency of the various methods. The computation time for each method with a fixed sample size of $2000$ is presented in column five of Table~\ref{tab:samples_complexity}.   All experiments were run on a single 2.4 GHz core.

Finally, we evaluated the empirical Type I error for $\alpha=5\%$ and increasing $B$. Figure~\ref{fig:blobs_emp_type1} displays the empirical Type I error, where we note the location of the  $\gamma=0.5$ heuristic in Equation~\eqref{eq:blockSelectionHeuristic}. For the user-chosen kernel ($\sigma=1$, Figure~\ref{fig:blobs_emp_type1a}), the number of Type I errors closely matches the targeted test level.  When median heuristic is used, however, the test is overly conservative, and makes fewer  Type I errors than required (Figure~\ref{fig:blobs_emp_type1b}). This  indicates that for this choice of $\sigma$, we are not in the asymptotic regime, and our Gaussian null distribution approximation is inaccurate. 
  Kernel selection via the strategy of \cite{GrettonSSSBPF2011} alleviates this
 problem (Figure~\ref{fig:blobs_emp_type1c}). 
This setting coincides with a block size substantially larger than 2 ($\MMD_l$), and therefore 
 achieves lower Type II errors while retaining the targeted  Type I error.

\vspace{-0.3cm}
\subsection{Musical experiments}\label{sec:MusicExperiment}
\begin{figure}
  \centering
  \subfigure[]{\label{fig:songs_emp_type1a}
    \includegraphics[width=0.31\textwidth]{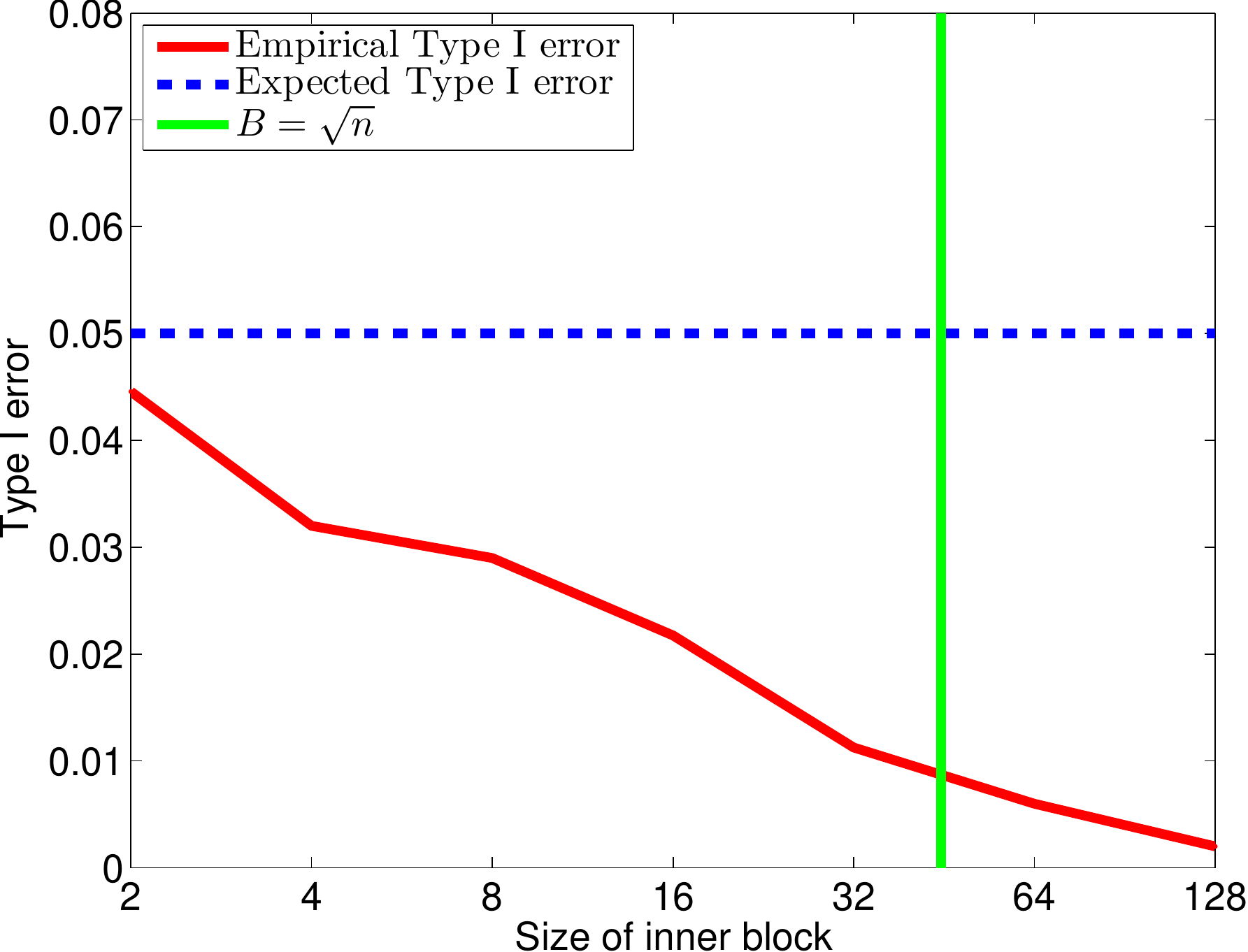}
  }
  \subfigure[]{\label{fig:songs_emp_type1b}
    \includegraphics[width=0.31\textwidth]{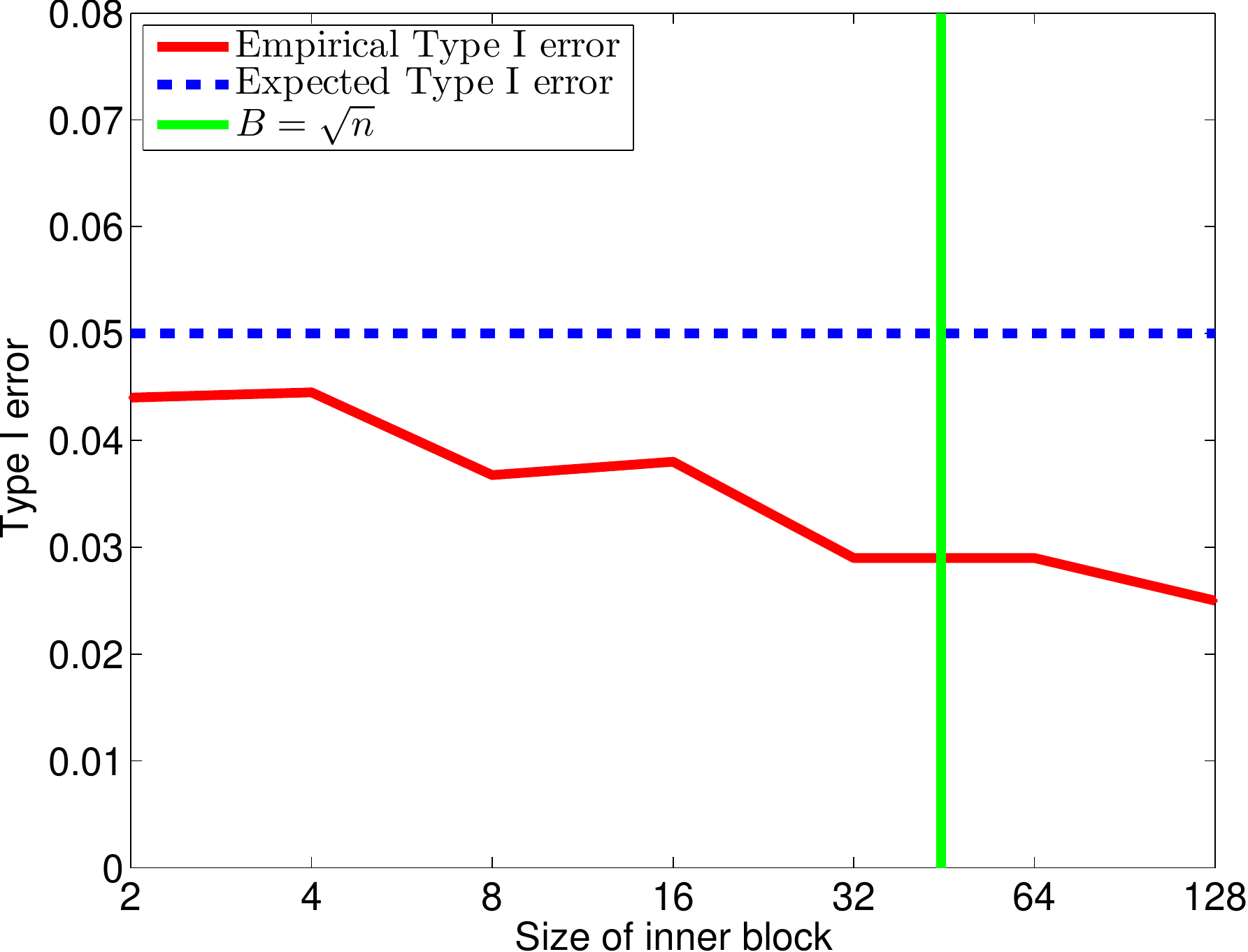}
  }
  \subfigure[]{\label{fig:songs_emp_type1c}
    \includegraphics[width=0.31\textwidth]{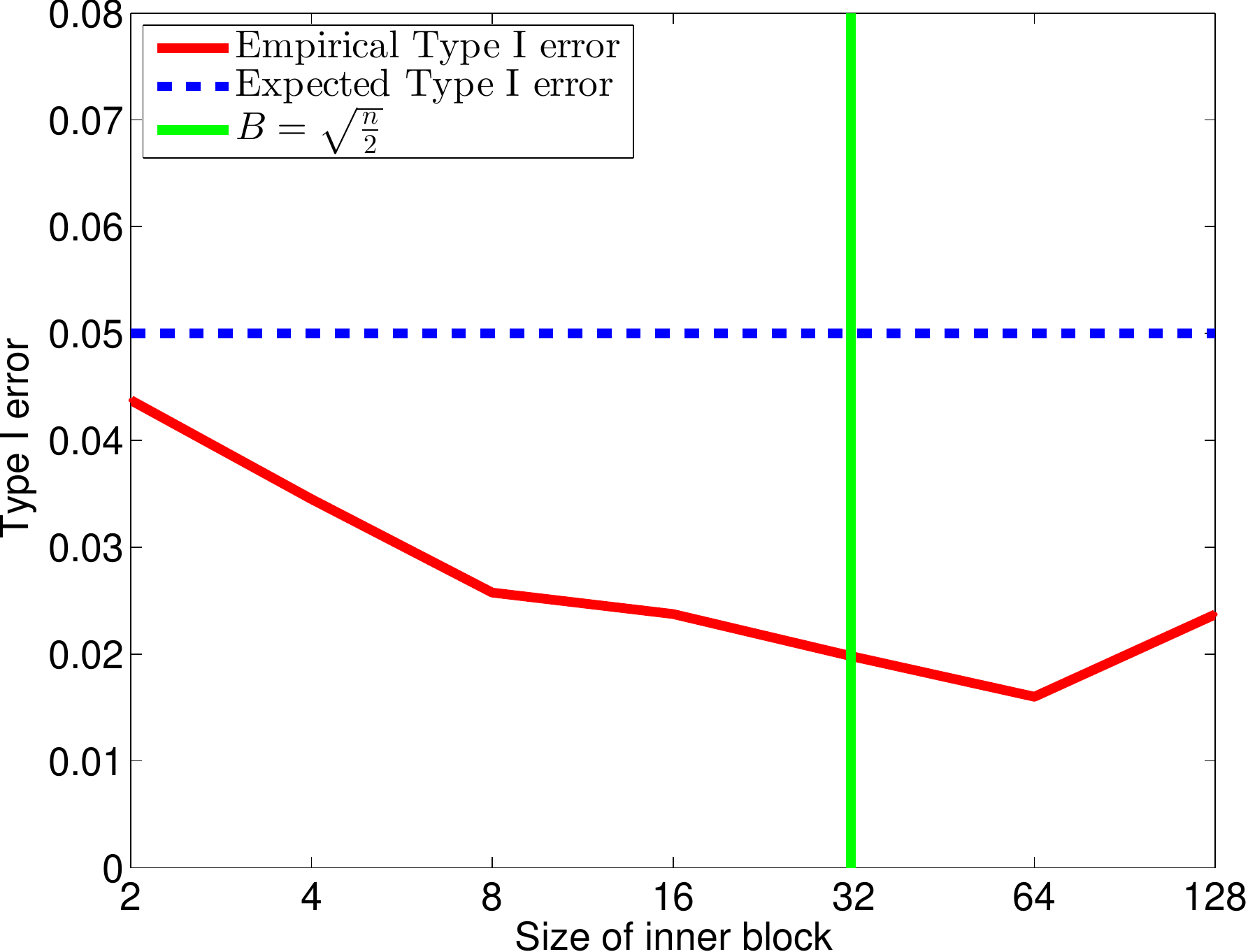}
  }  
  \caption{Empirical Type I error rate for $\alpha = 5\%$ on the music data (Section~\ref{sec:MusicExperiment}). (a) A single kernel test with $\sigma = 1$, (b) A single kernel test with $\sigma =$ median, and (c) for multiple kernels. Error bars are not visible at this scale.  The results broadly follow the trend visible from the synthetic experiments.}
  \label{fig:songs_emp_type1}
\end{figure}

\vspace{-0.2cm}
In this set of experiments, two amplitude modulated Rammstein songs were compared (\emph{Sehnsucht} vs.\ \emph{Engel}, from the album \emph{Sehnsucht}).
  Following the experimental setting in~\cite[Section 5]{GrettonSSSBPF2011}, samples from $P$ and $Q$ were extracts from AM signals of time duration $8.3 \times 10^{-3}$ seconds in the original audio.  Feature extraction was identical to~\cite{GrettonSSSBPF2011}, except that the amplitude scaling parameter was set to $0.3$ instead of $0.5$.  As the feature vector had size $1000$ we set the block size $B = \left\lceil \sqrt{1000} \right\rceil = 32$.  Table~\ref{tab:music_expr} summarizes the empirical Type I and Type II errors over $1000$ repetitions, and the average computation times.  Figure~\ref{fig:songs_emp_type1} shows the average number of Type I errors as a function of $B$: in this case, all kernel selection strategies result in conservative tests  (lower Type I error than required), indicating that more samples are needed to reach the asymptotic regime. Figure~\ref{fig:EmpiricalDistsLargeSmallK} shows the empirical $\mathcal{H}_0$ and $\mathcal{H}_A$ distributions for different $B$.

\begin{table*}\centering
\ra{0.9}
\resizebox{\textwidth}{!}{
\begin{tabular}{@{}cccccc@{}}\toprule

\multirow{2}{*}{Method} & \multirow{2}{60pt}{\centering Kernel parameters} & \multirow{2}{65pt}{\centering Additional parameters} & \multirow{2}{*}{Type I error} & \multirow{2}{*}{Type II error} & \multirow{2}{60pt}{\centering Computational time (s)}  \\

&&&&& \\
\cmidrule{1-6}
\multirow{6}{*}{$B$-test} & \multirow{2}{65pt}{\centering $\sigma = 1$} & $B = 2$& 0.038 & 0.927 & 0.039\\
 & & $B = \sqrt{n}$ & 0.006 & 0.597 & 1.276 \\
& \multirow{2}{65pt}{\centering $\sigma = \text{median}$} & $B = 2$& 0.043 & 0.786 & 0.047 \\
 & & $B = \sqrt{n}$ & 0.026 & 0 & 1.259 \\
 & \multirow{2}{65pt}{\centering multiple kernels} & $B = 2$& 0.0481 & 0.867 & 0.607 \\
 & & $B = \sqrt{\frac{n}{2}} $ & 0.025 & 0.012 & 18.285 \\
\cmidrule{1-6}
 Gram matrix spectrum & \multirow{2}{65pt}{\centering $\sigma = 1$} & \multirow{4}{65pt}{\centering $B = 2000$} & 0 & 0 & 160.1356 \\
 Bootstrap & & & 0.01 & 0 & 121.2570 \\
\cmidrule{1-2}\cmidrule{4-6}
 Gram matrix spectrum & \multirow{2}{65pt}{\centering $\sigma = \text{median}$} & & 0 & 0 & 286.8649 \\
 Bootstrap & & & 0.01 & 0 & 122.8297 \\
\bottomrule

\end{tabular}
}
    \caption{A comparison of consistent tests on the music experiment described in Section~\ref{sec:MusicExperiment}.  Here computation time is reported for the test achieving the stated error rates.}
    \label{tab:music_expr}
\end{table*}

\vspace{-0.3cm}
\section{Discussion}
\vspace{-0.3cm}

We have presented experimental results  both on a difficult synthetic problem, and on real-world data from amplitude modulated  audio recordings.
%
%
The results show that the $B$-test has a much better sample complexity than $\MMD_l$ over all tested kernel selection strategies. Moreover, it is an order of magnitude faster than any test that consistently estimates the null distribution for $\MMD_u$ (i.e., the  Gram matrix eigenspectrum and bootstrap estimates): these estimates are impractical at large sample sizes, due to their computational complexity.  Additionally, the $B$-test remains statistically consistent, with the best convergence rates achieved for large $B$.  The $B$-test  combines the best features of $\MMD_l$ and $\MMD_u$ based two-sample tests: consistency, high statistical efficiency, and high computational efficiency.

A number of further interesting experimental trends may be seen in these results.
First, 
we have observed that the empirical  Type I error rate is often conservative, and is less than the $5\%$ targeted by the threshold based on a Gaussian null distribution assumption (Figures~\ref{fig:blobs_emp_type1} and \ref{fig:songs_emp_type1}).  
In spite of this conservatism, the Type II performance remains strong (Tables~\ref{tab:samples_complexity} and~\ref{tab:music_expr}), as the gains in statistical power of the $B$-tests  improve  the testing performance (cf.\ Figure~\ref{fig:EmpiricalDistsLargeSmallK}).
Equation~\eqref{eq:ConvergenceHA} implies that the size of $B$ does not influence the asymptotic  variance under $\mathcal{H}_A$, however we observe 
in Figure~\ref{fig:EmpiricalDistsLargeSmallK}  that
the empirical variance of $\mathcal{H}_A$
 drops with larger $B$. This is because, for these $P$ and $Q$ and small $B$, the null and alternative distributions have considerable overlap. Hence, given the distributions are effectively indistinguishable at these sample sizes $n$, the variance of the alternative distribution as a function of $B$ behaves more like that of $\mathcal{H}_0$ (cf.\ Equation~\eqref{eq:ConvergenceH0}). This effect will vanish as $n$ grows.


Finally, \cite{HoShieh06} propose an alternative approach for U-statistic based testing  when the degree of degeneracy is known: a new U-statistic (the TU-statistic) is written in terms of products of centred U-statistics computed on the individual blocks, and a test is formulated using this TU-statistic. Ho and Shieh show that a TU-statistic based test can be asymptotically more powerful than a test using a single  U-statistic on the whole sample, when the latter is degenerate under $\mathcal{H}_0$, and nondegenerate under  $\mathcal{H}_A$. It is of interest to apply this technique to MMD-based two-sample testing.

\small
{\bf Acknowledgments} We thank Mladen Kolar for helpful discussions. This work is partially funded by ERC Grant 259112, and by the Royal Academy of Engineering through the Newton Alumni Scheme.
\normalsize

\bibliographystyle{plain}
\bibliography{bibliography}

\end{document}